\documentclass[conference]{IEEEtran}
\IEEEoverridecommandlockouts
\usepackage[
singlelinecheck=false 
]{caption}
\usepackage{cite}
\usepackage{amsmath,amssymb,amsfonts}
\usepackage{algorithmic}
\usepackage{graphicx}
\usepackage{textcomp}
\usepackage{xcolor}
\usepackage{booktabs}
\usepackage{subcaption}
\usepackage{cite}
\usepackage[section]{placeins}
\def\BibTeX{{\rm B\kern-.05em{\sc i\kern-.025em b}\kern-.08em
    T\kern-.1667em\lower.7ex\hbox{E}\kern-.125emX}}
\begin{document}
\title{LCA-Net: Light Convolutional Autoencoder for Image Dehazing}

\author{\IEEEauthorblockN{ Pavan A}
\IEEEauthorblockA{\textit{CSE PESU } \\
\textit{Bangalore, India}\\
\textit{pavana@pesu.pes.edu}\\
 \\
}
\and
\IEEEauthorblockN{ Adithya Bennur}
\IEEEauthorblockA{\textit{CSE PESU} \\
\textit{Bangalore, India }\\
\textit{aditya.bennur@gmail.com}\\
\\
}
\and
\IEEEauthorblockN{ Mohit Gaggar}
\IEEEauthorblockA{\textit{CSE PESU} \\
\textit{Bangalore, India }\\
\textit{gaggarmohit@gmail.com}\\
\\
}
\and
\IEEEauthorblockN{ Shylaja S S}
\IEEEauthorblockA{\textit{CSE PESU} \\
\textit{Bangalore, India }\\
\textit{shylaja.sharath@pes.edu}\\
\\
}

}

\maketitle

\begin{abstract}
Image dehazing is a crucial image pre-processing task aimed at removing the incoherent noise generated by haze to improve the visual appeal of the image. The existing models use sophisticated networks and custom loss functions which are computationally inefficient and requires heavy hardware to run. Time is of the essence in image pre-processing since real time outputs can be obtained instantly. To overcome these problems, our proposed generic model uses a very light convolutional encoder-decoder network which does not depend on any atmospheric models. The network complexity-image quality trade off is handled well in this neural network and the performance of this network is not limited by low-spec systems. This network achieves optimum dehazing performance at a much faster rate, on several standard datasets, comparable to the state-of-the-art methods in terms of image quality.

\end{abstract}


\section{Introduction}
Haze is an atmospheric phenomenon wherein the clarity of the sky is  obscured due to the presence of smoke, dust and other dry particulate matter. This causes light to scatter and thus, deteriorates the image quality which in turn may affect the performance of several computer vision applications such as detection, classification and tracking.
The atmospheric scattering model combines the haze-free
input with the global atmospheric illumination, blended together with a transmission coefficient that models the haze
effects. It is closely represented by the equation \cite{narasimhan2002vision}:
\begin{equation}
\textbf{I(x)=J(x)t(x) + A(x)(1 - t(x))}
\end{equation}

where x is the 2D pixel spatial location, I(x) is the observed
haze image,J(x) is the haze-free scene radiance to be recovered, while A(x) and t(x) are two critical parameters.
The first A(x), denotes the global atmospheric illumination, and t(x) is the transmission matrix.
\newline
Using the atmospheric model to build a neural network is also referred to as 'plug-in approach'. However, this approach may lack the features to detect intricate differences between a clean image and a hazy image. The viability of converting the problem of image reconstruction to a problem of estimation of A and t(x) is very low. Besides, we cannot relate hazy and clear images with ease, reason being the relation may be too compound to be explained just by an equation. Thus, the plug-in approach may not be able to generalize the equation over natural images.
Our model simply builds a clear image from the hazy image by applying pooling and convolutions and does not refer to the above atmospheric model or the transmission map of the image for the process of dehazing of the image.

Image Dehazing is used in several image analyzing applications and other applications such as tracking, control systems, and intelligent surveillance where good quality images or videos are crucial for obtaining accurate results and optimum performance.

\section{THE PROPOSED NETWORK}
We shall propose a convolutional autoencoder in this section which is light both in terms of the architecture as well as its speed in dehazing an image. As the name suggests, this neural network has no special consideration towards image dehazing in its design. The following subsections outlines overall architecture, important features, main building blocks and the loss function used of the LCA-net.

\subsection{    Network Architecture}\label{AA}
\begin{figure*}[ht]
\captionsetup{justification=centering}
\centerline{\includegraphics[width=\textwidth,height=4.5cm]{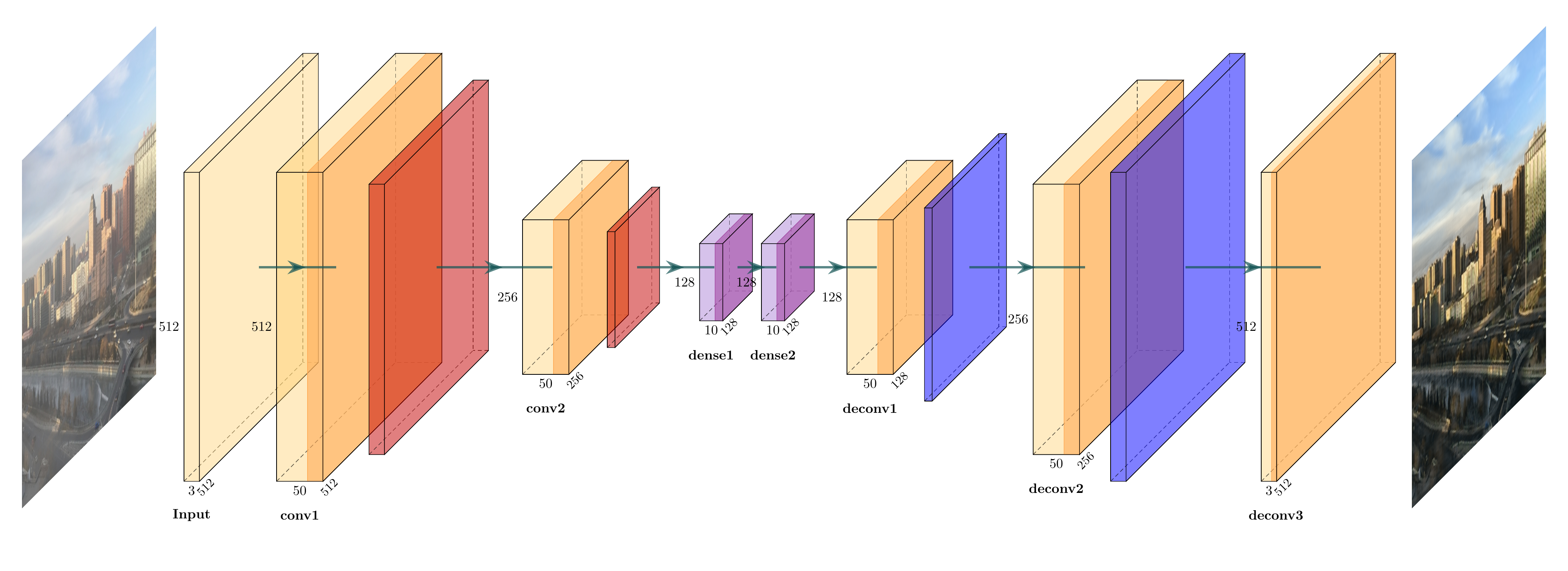}}
\caption{The architecture of LCA-Net}
\label{fig:lcanet}
\end{figure*}

Figure \ref{fig:lcanet} illustrates the overall architecture of LCA-Net. Our neural network consists of an encoder-decoder type of architecture. The input layer takes an image (512 x 512 x 3) which is padded and pushed to the first ReLU \cite{maas2013rectifier} activated convolutional layer consisting of 50 filters of size 3. This layer returns an output of original dimensions which acts as input for the average pooling layer which has a downsampling factor of 2, followed by the next convolutional layer where the cycle is repeated. This results in an encoded input of size 128 x 128 x 50.
This stage is referred to as 'bottleneck'. The encoded input is then passed through 2 completely connected ReLU \cite{maas2013rectifier} activated dense layers consisting of 10 neurons each. Dense layers add an interesting non-linearity property, and can model any mathematical function. Thus, they help in extracting essential features efficiently.
This output is passed to the decoder which consists of 2 pairs of deconvolutional and upsampling layers. The ReLU\cite{maas2013rectifier} activated deconvolutional layer consists of 50 filters of size 3 and the upsampling layer utilises a scale factor of 2.
The resulting output of the decoder has a dimension of 512 x 512 x 50. The final output layer is a deconvolutional layer consisting of 3 filters of size 3 which are linearly activated producing a dehazed image whose size is identical to the input image.

\subsection{    Loss function and Training setting}
\begin{figure}[ht]
\centering
\captionsetup{justification=centering}
\centerline{\includegraphics[width=75mm,scale=0.8]{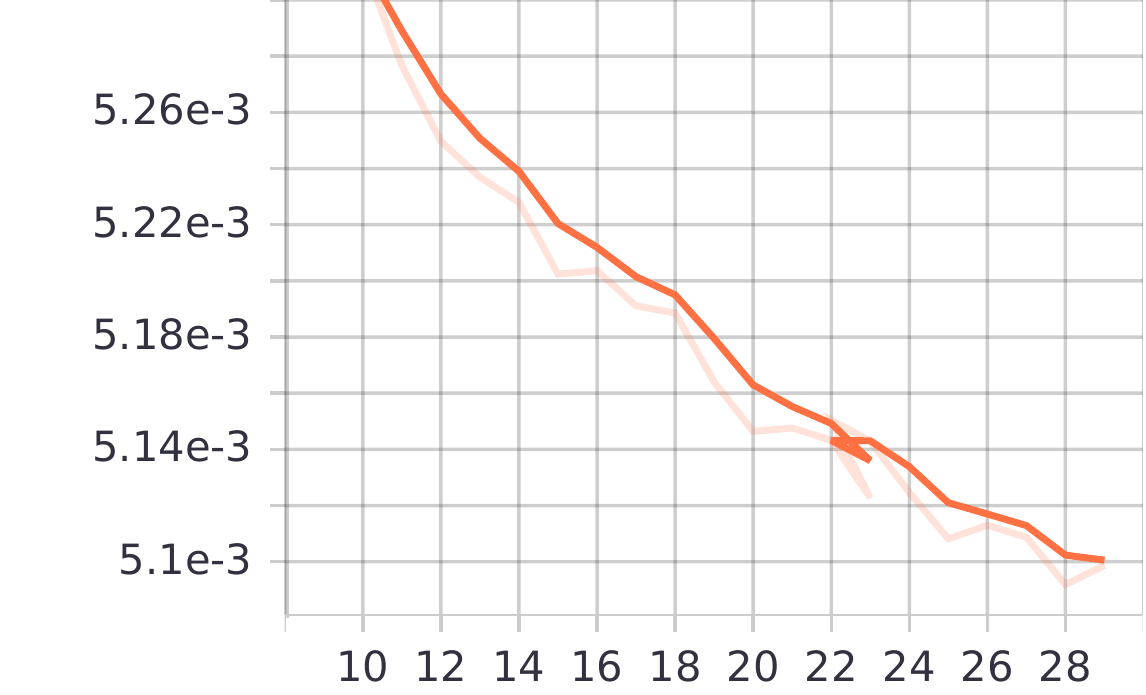}}
\caption{Epoch loss} 
\label{fig:loss}
\end{figure}

The loss function used in this model is mean squared error(MSE). There are high chances of encountering unexpected changes in intensity of pixel values. MSE helps average the variation of extreme values, supporting a smooth gradient descent making it an optimal loss function. 
\begin{equation}
L=\frac{1}{N}\sum_{x=1}^{N}\sum_{i=1}^{3}\left \| \hat{J}_{i}\left ( x \right )-J_{i}\left ( x \right ) \right \|^{2}
\end{equation}

where J\textsubscript{i}(x) is the intensity of the i\textsuperscript{th} color channel of pixel x in the output image (the ground truth), and N is the total number of pixels. Figure \ref{fig:loss} shows the loss function decreasing with increase in the number of epochs.
\newline
The optimizer used is ADAM \cite{kingma2014adam} in its default settings which are: learning rate = 0.001 , 
$ \beta_{1}=0.9 , \beta_{2}=0.999.$ 

\subsection{    Datasets}
Like most of the previous CNN based dehazing techniques our autoencoder is also trained on a custom dataset. Our dataset consists a total of 2061 clear images, we then select 35 varying haze levels on each image to constitute a total of 72135 haze images borrowed from the RESIDE dataset \cite{li2019benchmarking}.
\newline
There are 3 test sets borrowed from RESIDE-standard \cite{li2019benchmarking} dataset. The test sets, known as the Synthetic Objective Testing Set (SOTS-outdoor and SOTS-indoor) contain 500 synthetic outdoor and 500 synthetic indoor hazy images respectively. They are generated using images in NYU2 and are different from the training set images. The third test set is known as the Hybrid Subjective Testing Set (HSTS), which contains 10 synthetic outdoor and 10 real-world outdoor hazy images to evaluate qualitative visual performance.
\newline

\section{Benchmarking and Results}
The performance of the neural network is determined by evaluation metrics such as PSNR(Peak Signal to Noise Ratio) and SSIM(Structural Similarity Index). PSNR is essentially the ratio between maximum intensity and MSE. High PSNR value implicates low noise, the vice-versa is also true. 
SSIM as the name says measures the similarity between 2 images. It also improves on methods like PSNR and MSE.
\newline
\newline
Table \ref{tab:1}, Table \ref{tab:2} and Table \ref{tab:3} shows the PSNR and SSIM scores of our LCA-Net compared to nine other state-of-the-art methods \cite{5567108, 5459251, Meng_2013_ICCV, 10.1007/978-3-319-46475-6_36, Berman_2016_CVPR, 7539399, 10.1007/978-3-319-46475-6_10, Li_2017_ICCV, 8803478} on the HSTS, SOTS-indoor and SOTS-outdoor test sets respectively. The results of previous dehazing methods for Table \ref{tab:1}, Table \ref{tab:2} and Table \ref{tab:3} are from, \cite{li2019benchmarking} and \cite{8803478}. Nonetheless, since our model benchmarks on the same dataset, the comparison is fair.
\newline

\begin{table*}[hbt!]
\centering
\captionsetup{justification=centering}
\renewcommand{\arraystretch}{2}
\begin{tabular}{|c|c|c|c|c|c|c|c|c|c|c|}
\hline
     & DCP \cite{5567108}& FVR \cite{5459251} & BCCR \cite{Meng_2013_ICCV}   & GRM \cite{10.1007/978-3-319-46475-6_36}    & NLD \cite{Berman_2016_CVPR} & DehazeNet \cite{7539399}       & MSCNN \cite{10.1007/978-3-319-46475-6_10}  & AOD-Net \cite{Li_2017_ICCV} & CAE \cite{8803478} & LCA-Net         \\ \hline
PSNR & 14.84  & 14.48      & 15.08  & 18.54  & 18.92      & 24.48           & 18.64  & 20.55   & 20.08      & \textbf{24.734} \\ \hline
SSIM & 0.7609 & 0.7624     & 0.7382 & 0.8184 & 0.7411     & \textbf{0.9153} & 0.8168 & 0.8973  & 0.8169     & 0.8951          \\ \hline
Time(s) & 1.62 & 6.79 & 3.85 & 83.96 & 9.89 & 2.51 & 2.60 & 0.65 & 1.13 & \textbf{0.3546} \\
\hline
\end{tabular}

\caption{Dehaze results on synthetic hazy images from HSTS}
\label{tab:1}
 \vspace{7mm}

\centering
\renewcommand{\arraystretch}{2}
\captionsetup{justification=centering}
\begin{tabular}{|c|c|c|c|c|c|c|c|c|c|c|}
\hline
     & DCP \cite{5567108}& FVR \cite{5459251} & BCCR \cite{Meng_2013_ICCV}   & GRM \cite{10.1007/978-3-319-46475-6_36}    & NLD \cite{Berman_2016_CVPR} & DehazeNet \cite{7539399}       & MSCNN \cite{10.1007/978-3-319-46475-6_10}  & AOD-Net \cite{Li_2017_ICCV} & CAE \cite{8803478} & LCA-Net \\ \hline
PSNR & 16.62  & 15.72  & 16.88  & 18.86  & 17.29  & 21.14     & 17.57  & 19.06   & \textbf{24.56}  & 18.23   \\ \hline
SSIM & 0.8179 & 0.7483 & 0.7913 & 0.8553 & 0.7489 & 0.8472    & 0.8102 & 0.8504  & \textbf{0.9126} & 0.7808  \\ \hline
\end{tabular}
\caption{Dehaze results on Indoor hazy images from SOTS}
\label{tab:2}
\vspace{7mm}
\centering
\renewcommand{\arraystretch}{2}
\captionsetup{justification=centering}
\begin{tabular}{|c|c|c|c|c|c|c|c|c|c|}
\hline
 & DCP \cite{5567108}& FVR \cite{5459251} & BCCR \cite{Meng_2013_ICCV}   & GRM \cite{10.1007/978-3-319-46475-6_36}    & NLD \cite{Berman_2016_CVPR} & DehazeNet \cite{7539399}       & MSCNN \cite{10.1007/978-3-319-46475-6_10}  & AOD-Net \cite{Li_2017_ICCV} & LCA-Net \\ \hline
PSNR & 18.54 & 16.61 & 17.71 & 20.77 & 19.52 & \textbf{26.84} & 21.73 & 24.08 & 23.37 \\ \hline
SSIM & 0.71 & 0.7236 & 0.7409 & 0.7617 & 0.7328 & 0.8264 & 0.8313 & 0.8726 & \textbf{0.8763} \\ \hline
\end{tabular}
\caption{Dehaze results on Outdoor hazy images from SOTS}
\label{tab:3}

\end{table*}

\begin{figure*}[ht!]
     \centering
    \captionsetup{justification=centering}
    \begin{subfigure}[b]{0.15\textwidth}
        \includegraphics[width=25mm]{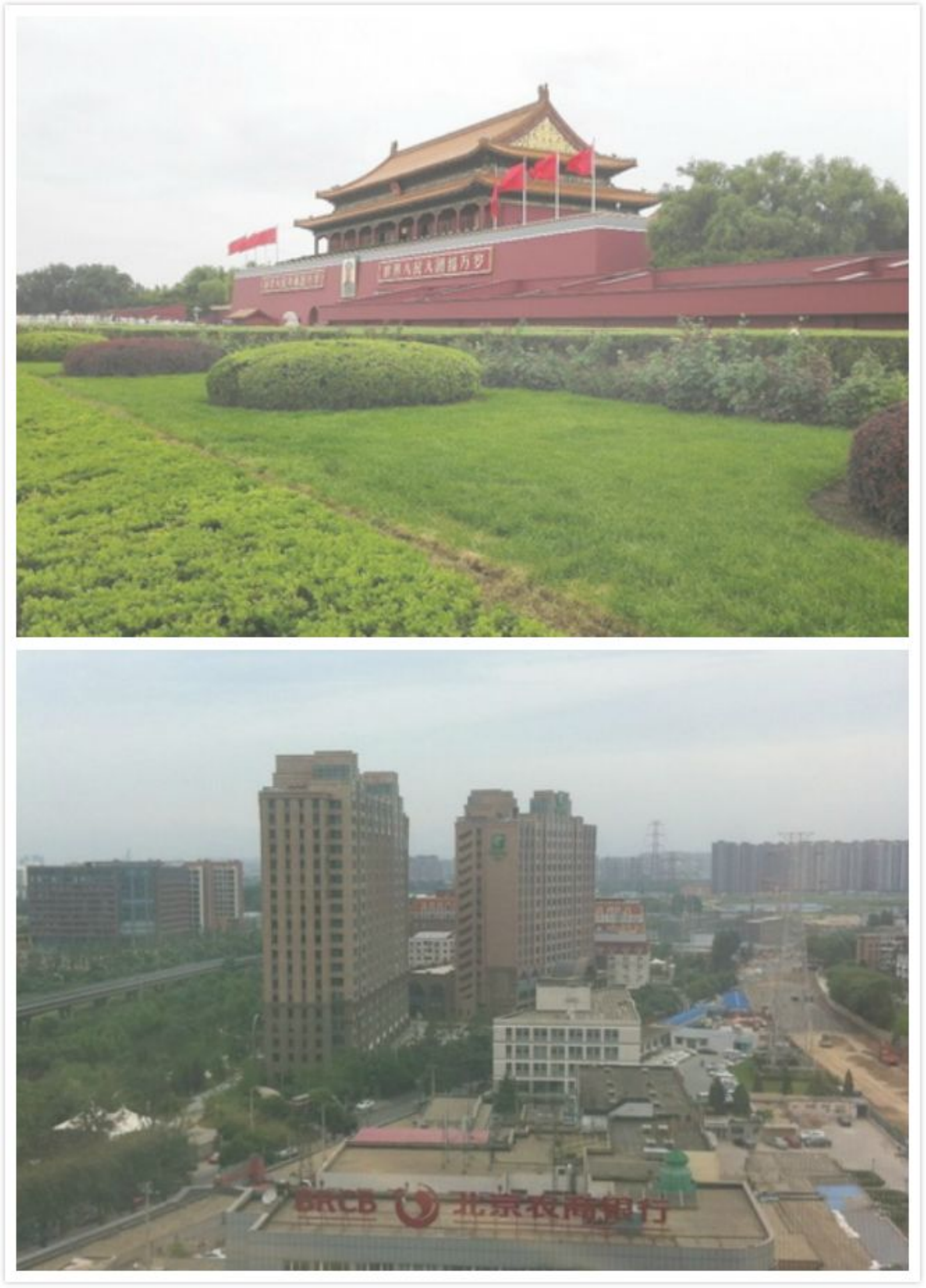}
        \caption{hazy}
        \label{fig:hazy image}
    \end{subfigure}
    \begin{subfigure}[b]{0.15\textwidth}
        \includegraphics[width=25mm]{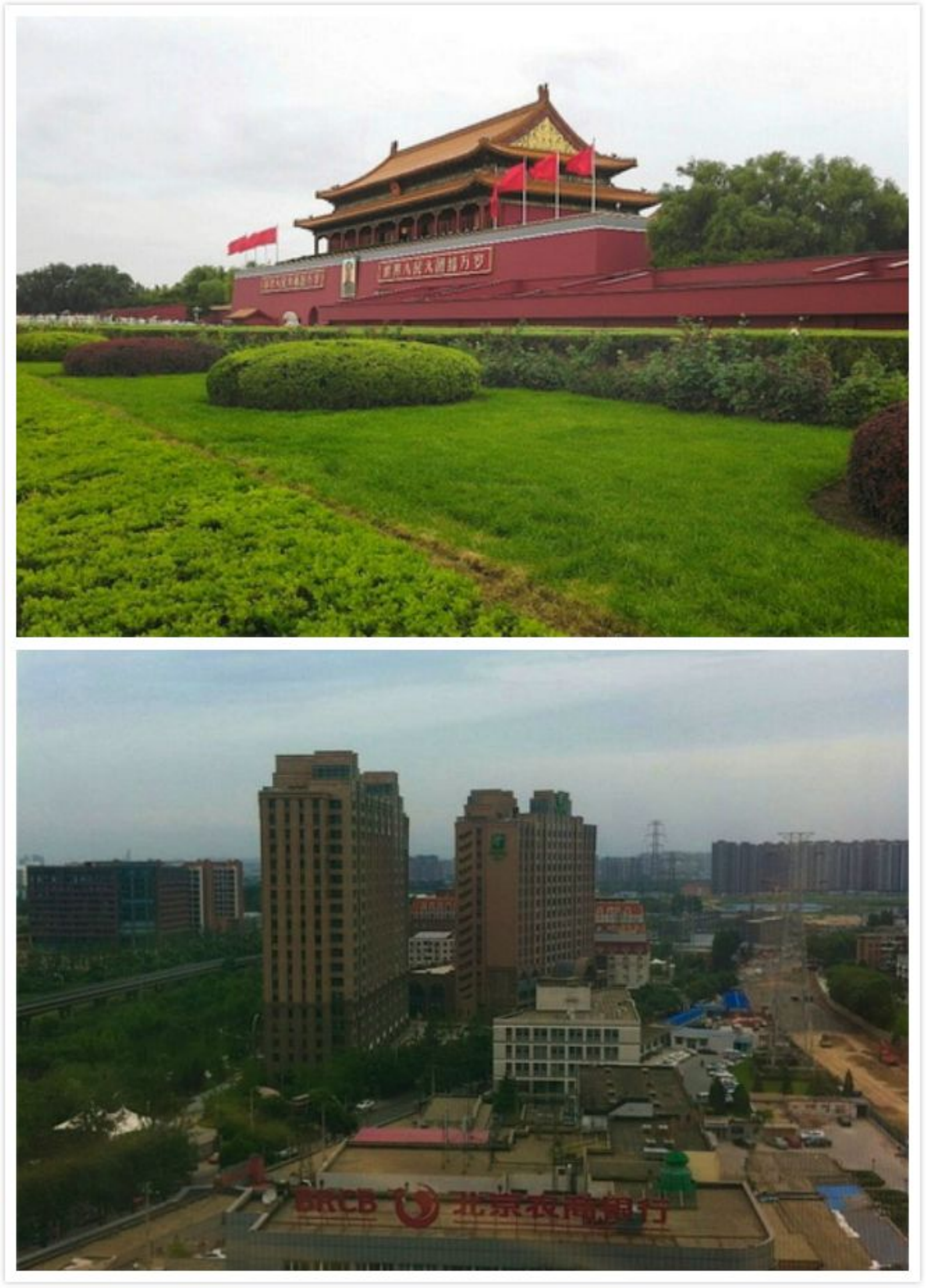}
        \caption{AOD-Net}
        \label{fig:AOD-net}
    \end{subfigure}
    \begin{subfigure}[b]{0.15\textwidth}
        \includegraphics[width=25mm]{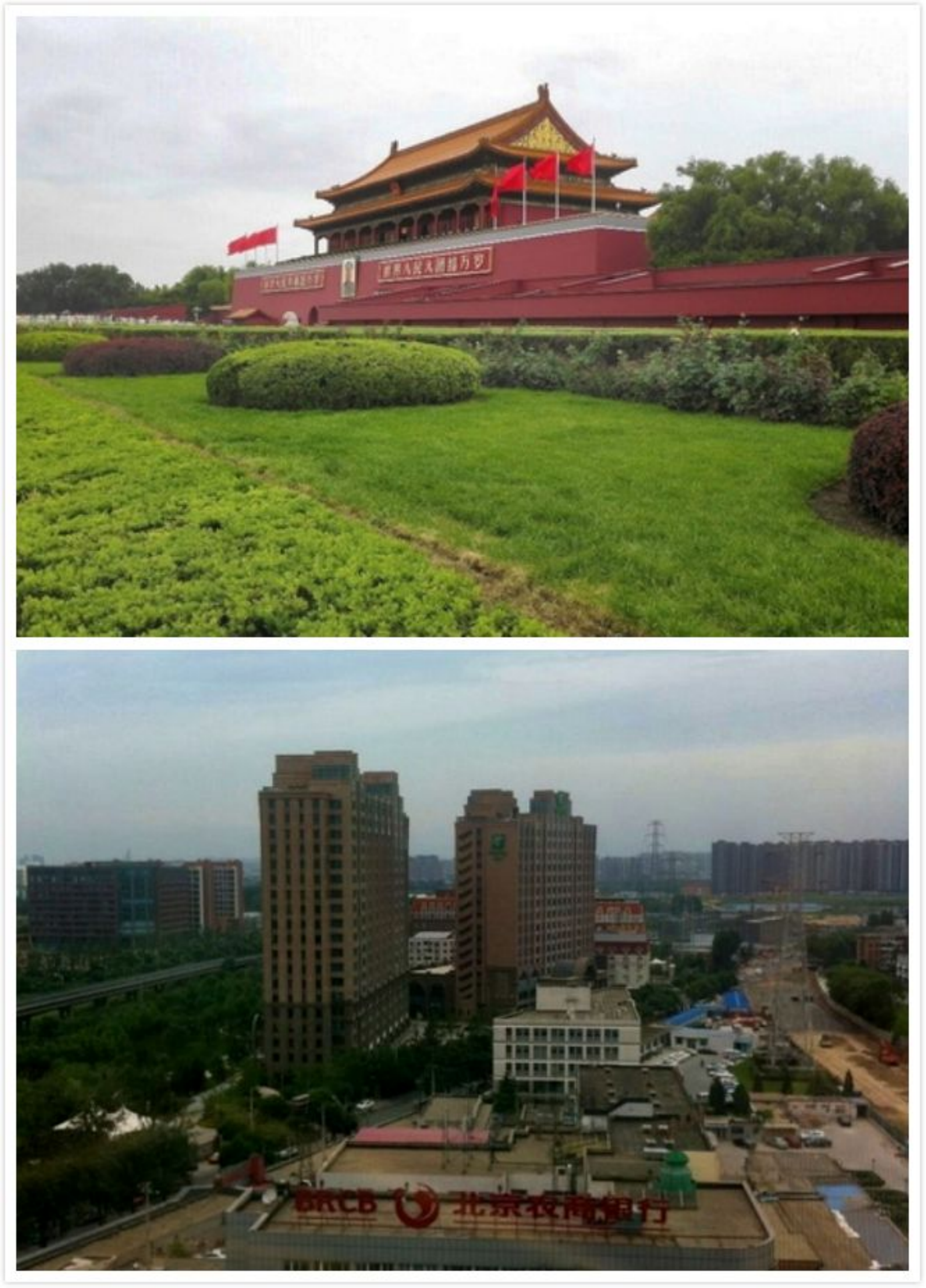}
        \caption{dehazeNet} 
        \label{fig:Dehaze-net}
    \end{subfigure} 
    \begin{subfigure}[b]{0.15\textwidth}
        \includegraphics[width=25mm]{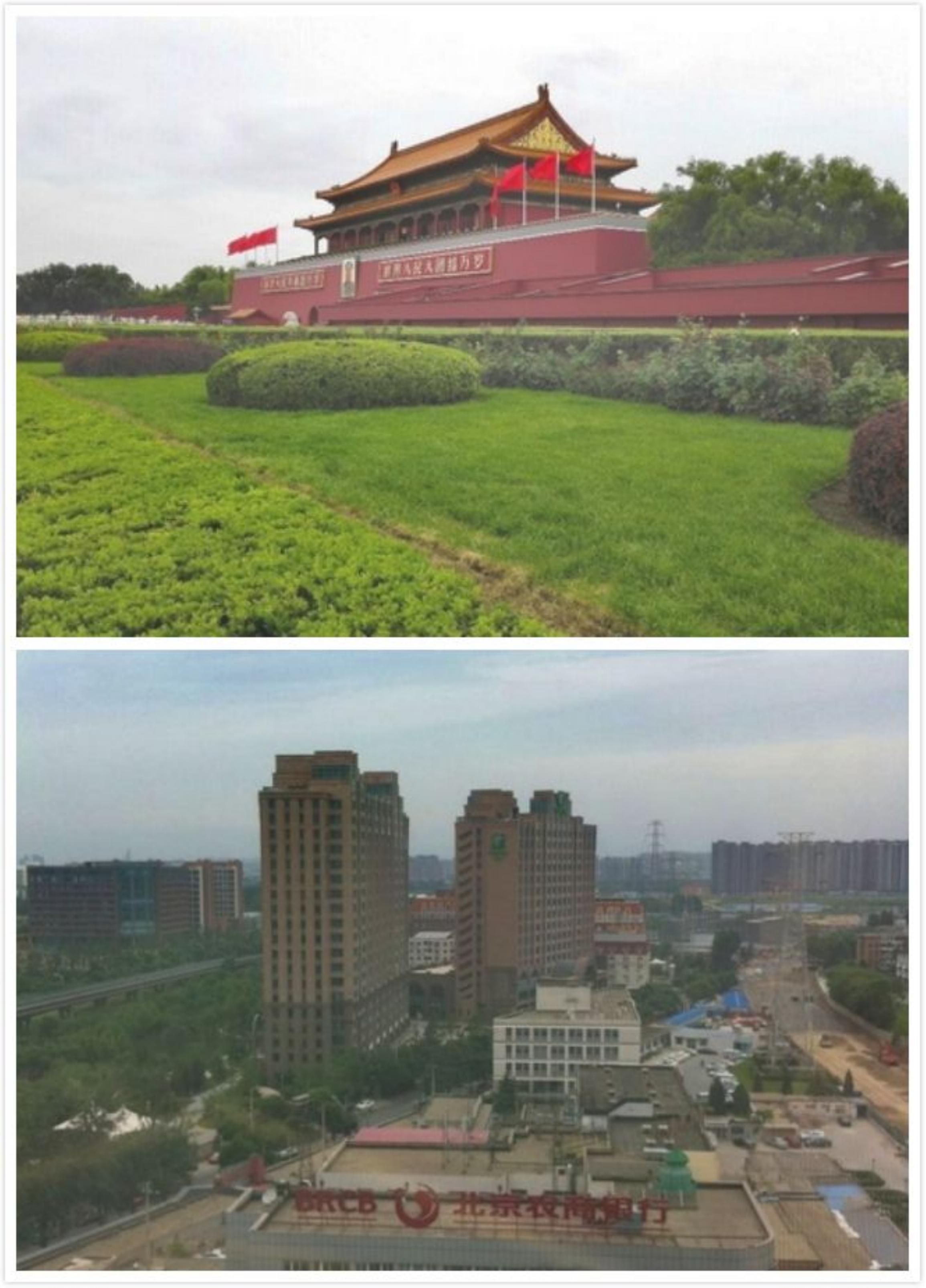}
        \caption{MSCNN}
        \label{fig:MSCNN}
    \end{subfigure}
     \begin{subfigure}[b]{0.15\textwidth}
        \includegraphics[width=25mm]{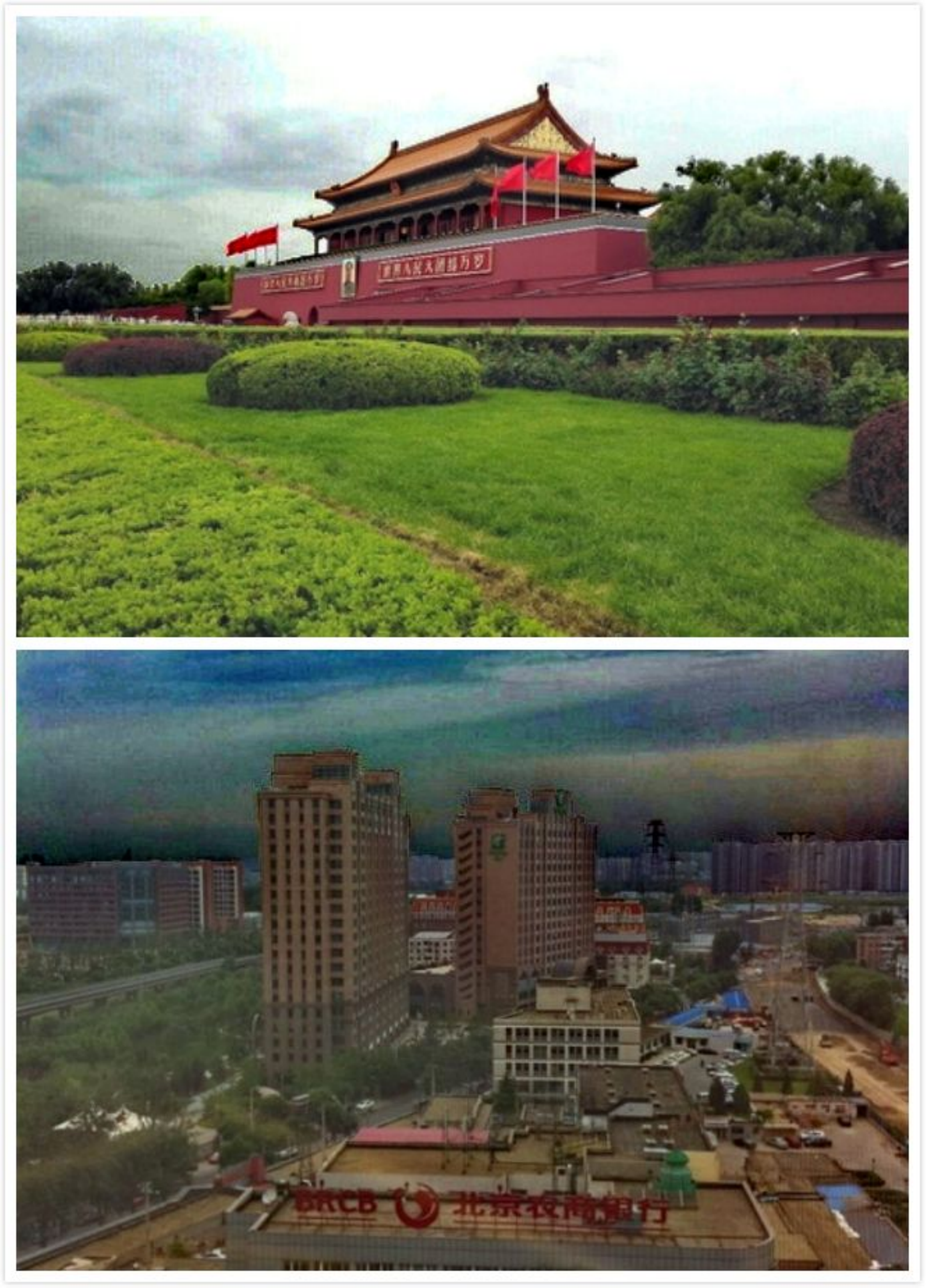}
        \caption{NLD}
        \label{fig:NLD}
    \end{subfigure}
     \begin{subfigure}[b]{0.15\textwidth}
        \includegraphics[width=25mm]{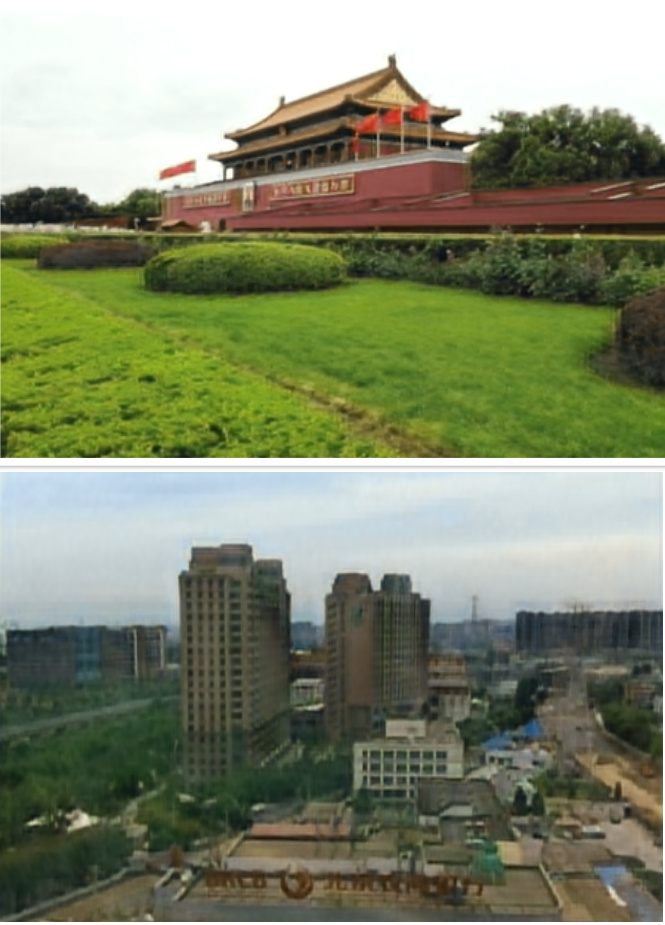}
        \caption{LCA-Net}
        \label{fig:LCA-net}
    \end{subfigure}
\caption{Comparison of various model benchmarks}
\label{fig:benchmarkfig}

\end{figure*}

Table \ref{tab:1} shows the PSNR, SSIM scores and average time taken to dehaze an image. Our model when compared to other state-of-the-art method \cite{5567108, 5459251, Meng_2013_ICCV, 10.1007/978-3-319-46475-6_36, Berman_2016_CVPR, 7539399, 10.1007/978-3-319-46475-6_10, Li_2017_ICCV, 8803478} comes in first and the SSIM only comes second to DehazeNet \cite{7539399} in outdoor HSTS. Due to the light nature of our model it outperforms all other models in the run-time per image, making it one of the fastest methods for image dehazing. 
\newline
\newline
Table \ref{tab:2} shows the dehaze results on SOTS-indoor dataset. Our model performs inadequately when bench-marked on indoor SOTS and low-light images as they were trained on outdoor images. This phenomenon can also be explained due to the drastically varying feature vectors of indoor and outdoor images. Deficient performance on low-light images is also explained by the fact that low-light images have contrasted feature vectors. Figure \ref{fig:benchmarkfig} illustrates the comparison of different models' performances on standard hazy images from the RESIDE dataset \cite{li2019benchmarking}.
\newline
\newline
Table \ref{tab:3} shows the PSNR and SSIM scores of models \cite{5567108, 5459251, Meng_2013_ICCV, 10.1007/978-3-319-46475-6_36, Berman_2016_CVPR, 7539399, 10.1007/978-3-319-46475-6_10, Li_2017_ICCV} on SOTS-outdoor images. Our model comes third behind DehazeNet \cite{7539399} and AOD-Net \cite{Li_2017_ICCV} in PSNR and first in the SSIM metric. Therefore, our model is efficient in dehazing outdoor real world and synthetic images. 
\section{Conclusion and Future Work}This paper presented a light convolutional autoencoder which quickly helps remove haze in images. Further it performs better than other state-of-the-art image dehazing methods with regard to PSNR and the time taken to dehaze an image. The image quality is comparable to the state of the art methods. Since it is a light weighted network, it is also compatible with low-spec computers. The downside of this network is its inefficient performance on low-light images and dense-haze images and our future work would be centered on eliminating this flaw to provide fine performance on all categories of images. 
\nocite{*}
\bibliographystyle{unsrt}
\bibliography{BibFile}

\end{document}